%% file: root.tex
\documentclass[letterpaper, 10 pt, conference]{ieeeconf}
\IEEEoverridecommandlockouts
\usepackage{cite}
\usepackage{tocloft}
\usepackage{graphicx}
\usepackage{multirow}
\usepackage{makecell}
\usepackage{amssymb}
\usepackage{amsmath}
\usepackage{mathtools}
\usepackage{hyperref}
\usepackage{booktabs}
\usepackage{subcaption}

\input{macro}

\title{\LARGE \bf
Clothes Grasping and Unfolding Based on RGB-D \\ Semantic Segmentation
}

\author{Xingyu Zhu$^{1,2}$, Xin Wang$^{1,2}$, Jonathan Freer$^{3}$, Hyung Jin Chang$^{3}$, Yixing Gao$^{1,2,*}$
\thanks{$^{1}$ School of Artificial Intelligence, Jilin University, China.}
\thanks{$^{2}$ Engineering Research Center of Knowledge-Driven Human-Machine Intelligence, Ministry of Education, China.}
\thanks{$^{3}$ School of Computer Science, University of Birmingham, UK.}
\thanks{*Corresponding author. Email: 
\href{mailto:gaoyixing@jlu.edu.cn}{gaoyixing@jlu.edu.cn}
}
}

\begin{document}

\maketitle
\thispagestyle{empty}
\pagestyle{empty}

\begin{abstract}
Clothes grasping and unfolding is a core step in robotic-assisted dressing. Most existing works leverage depth images of clothes to train a deep learning-based model to recognize suitable grasping points. These methods often utilize physics engines to synthesize depth images to reduce the cost of real labeled data collection. However, the natural domain gap between synthetic and real images often leads to poor performance of these methods on real data. Furthermore, these approaches often struggle in scenarios where grasping points are occluded by the clothing item itself. To address the above challenges, we propose a novel Bi-directional Fractal Cross Fusion Network (BiFCNet) for semantic segmentation, enabling recognition of graspable regions in order to provide more possibilities for grasping. Instead of using depth images only, we also utilize RGB images with rich color features as input to our network in which the Fractal Cross Fusion (FCF) module fuses RGB and depth data by considering global complex features based on fractal geometry. To reduce the cost of real data collection, we further propose a data augmentation method based on an adversarial strategy, in which the color and geometric transformations simultaneously process RGB and depth data while maintaining the label correspondence. Finally, we present a pipeline for clothes grasping and unfolding from the perspective of semantic segmentation, through the addition of a strategy for grasp point selection from segmentation regions based on clothing flatness measures, while taking into account the grasping direction. We evaluate our BiFCNet on the public dataset NYUDv2 and obtained comparable performance to current state-of-the-art models. We also deploy our model on a Baxter robot, running extensive grasping and unfolding experiments as part of our ablation studies, achieving an 84\% success rate.

\end{abstract}
\section{INTRODUCTION}
In recent years, robot-assisted dressing has gained increased attention in the field of assistive robotics \cite{zhang2022learning,gao2015user,zhang2019probabilistic,gao2016iterative,zhang2020learning,saxena2019garment,gao2020user}. Zhang and Demiris propose a robot-assisted dressing pipeline for people with limited limb mobility, composed of clothes grasping and unfolding, robot navigation, and user upper-body dressing \cite{zhang2022learning}. This task is highly dependent on the robots ability to recognize grasping points, and many methods have been proposed to tackle this sub task. Most existing works apply deep learning methods for keypoints recognition leveraging depth images of the clothing. Due to collection of real labeled depth images being extremely time consuming, physics engines such as Maya and Blender are usually used for data enrichment \cite{ren2021grasp,saxena2019garment,zhang2020learning,zhang2022learning,bousmalis2018using}. However, a clear domain gap is present between the synthesized and real depth images, with the real depth images containing significant noise. Furthermore, a reliance on depth information only, results in existing methods struggling in situations where pre-defined points are self occluded by the garment itself.

\begin{figure}[t]
      \centering
      \includegraphics[width=7.5cm]{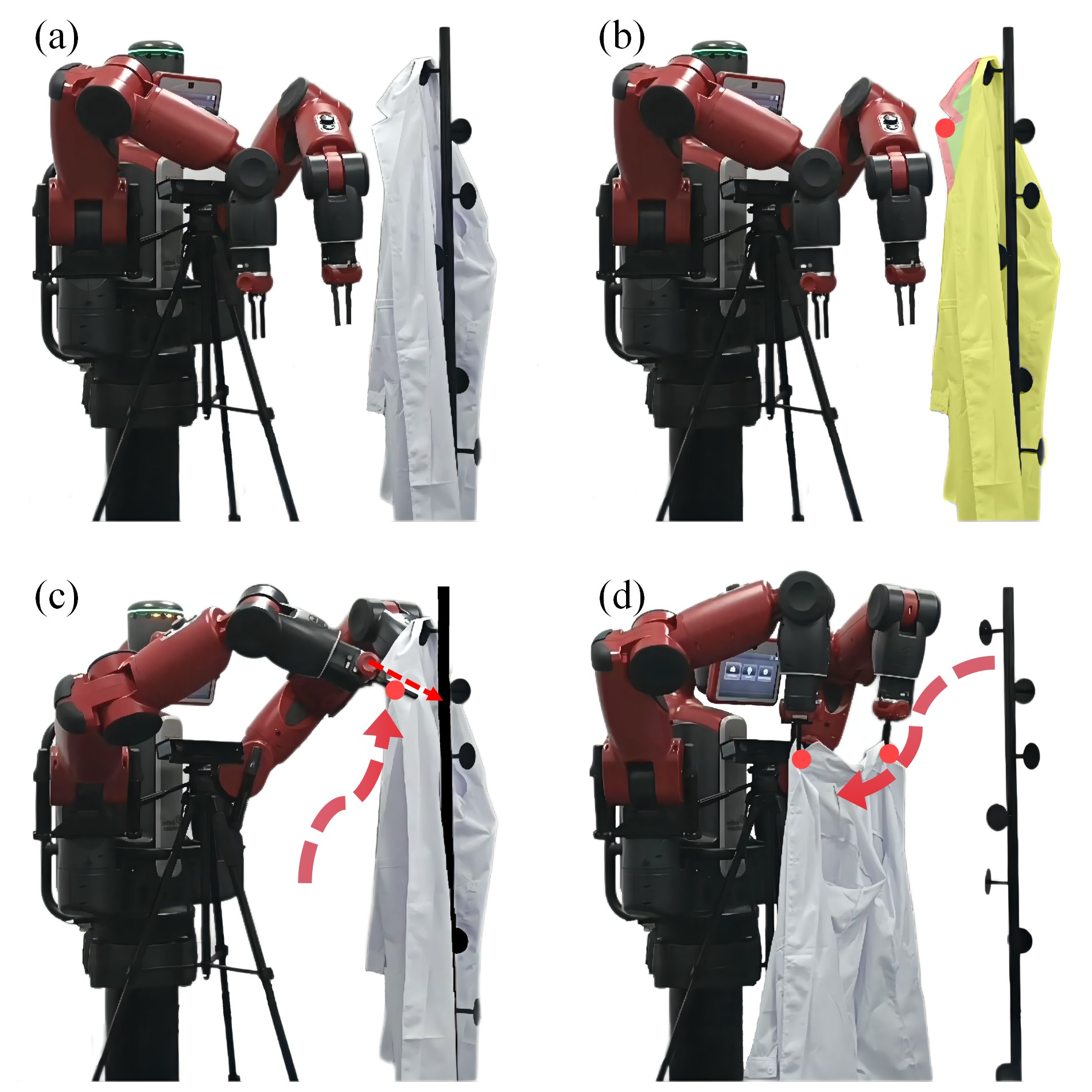}
      \caption{Clothes grasping and unfolding. (a) Experimental setup. (b) Clothes segmentation and calculation of grasping points. (c) Grasping operation. (d) Unfolding operation.}
      \label{f1}
      \vspace{-0.5cm}
\end{figure}
\begin{figure*}[!thpb]
      \centering
      \includegraphics[width=14cm]{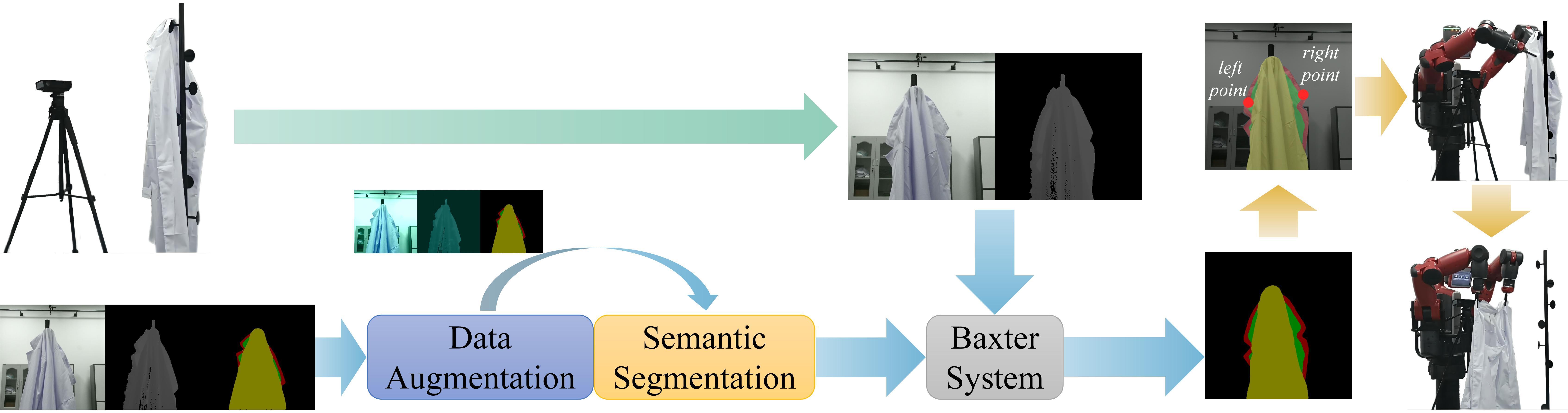}
      \caption{Overview of the RGB-D semantic segmentation pipeline for clothes grasping and unfolding. The item of clothing is segmented into 3 parts, inner collar, outer collar and the remaining area, for the selection of grasping points based on the inner and outer collar later. Given a collection of RGB images, depth images and corresponding ground-truth segmentation results, we propose a data augmentation method for data enrichment. Then, we train a semantic segmentation model on the clothing dataset using the proposed BiFCNet and deploy it on the Baxter robot. Given real-time RGB and depth images, the robot can distinguish different regions of the clothes, decide grasping points, perform the grasping and unfolding motion.}
      
      \label{f2}
      \vspace{-0.5cm}
\end{figure*}


In this paper, we propose a novel Bi-directional Fractal Cross Fusion Network (BiFCNet) for semantic segmentation and apply it in the scenario of clothes grasping and unfolding, as shown in Fig \ref{f1}. We employ semantic segmentation to recognize graspable regions on the clothing item. Unlike key point recognition, we find that segmentation provides rich information allowing for more accurate grasping point selection. As an addition to depth only methods, RGB data can also be leveraged for this task, allowing for greater recognition of the folded area of clothing item.

Recently, RGB-D semantic segmentation has been applied in a variety of complex scenes \cite{hazirbas2016fusenet,park2017rdfnet,zhang2019pattern,xiong2020variational,xing2020malleable,hua2019small,schwarz2018rgb}. In the majority of these works, two branches are used to extract features from RGB and depth images separately and later fuse the feature representations. These methods, however, mainly consider the fusion of local features and ignore the fusion of global features \cite{chen2020bi,seichter2021efficient,cao2021shapeconv,su2021deep,wang2022multimodal}. Global features are crucial for enabling the complex spatial information of deep features to assist the propagation of RGB semantic information, and fractal geometry has been shown to be an effective method for extracting globally complex features \cite{xu2021encoding}.
In our proposed BiFCNet, the Fractal Cross Fusion (FCF) module can realize the fusion of RGB data and depth data by considering global complex features based on fractal geometry.

Moreover, we propose an adversarial strategy-based data augmentation method for RGB-D semantic segmentation that can process both RGB and depth images based on color and geometric transformations while maintaining consistency with the image labels.
Data augmentation methods based on adversarial strategies have been proven to be a better choice for enhancing the generalization ability of models in the case of limited data \cite{cubuk2019autoaugment,suzuki2022teachaugment}. We notice that the type of clothes in current robot-assisted dressing research is quite limited \cite{zhang2022learning,gao2015user,zhang2019probabilistic,gao2016iterative,zhang2020learning}. Our proposed adversarial strategy-based data augmentation method directly addresses this, facilitating training with small amounts of labeled real data even when the type of clothing changes.

With the aforementioned proposed methods, we introduce an RGB-D semantic segmentation pipeline for clothes grasping and unfolding, as shown in Fig \ref{f2}. Based on the segmentation results of the clothes, we select the most suitable grasping points from the identified graspable regions by establishing a flatness measurement model for different regions of the item of clothing. The main contributions can be summarized as follows:  

1) We propose an RGB-D semantic segmentation model BiFCNet, in which the FCF module fuses RGB information and depth information based on fractal geometry by considering global complex features. 

2) We propose a data augmentation method based on an adversarial strategy, in which the color and geometric transformations can simultaneously process RGB and depth data while maintaining the label correspondence.

3) We propose an RGB-D semantic segmentation pipeline for clothes unfolding, through the combination of the aforementioned contributions with a strategy to select graspable points from segmented regions, based on clothing flatness measures while taking into account the grasping direction.

4) We evaluate BiFCNet on the NYUDv2 dataset, with mIoU and PA of 51.8\% and 77.9\% respectively, reaching a level comparable to the state-of-the-art methods.We also evaluate our method through an extensive ablation study focusing on real world robotic clothes grasping and unfolding.

\begin{figure*}[!thpb]
      \centering
      \includegraphics[width=15cm]{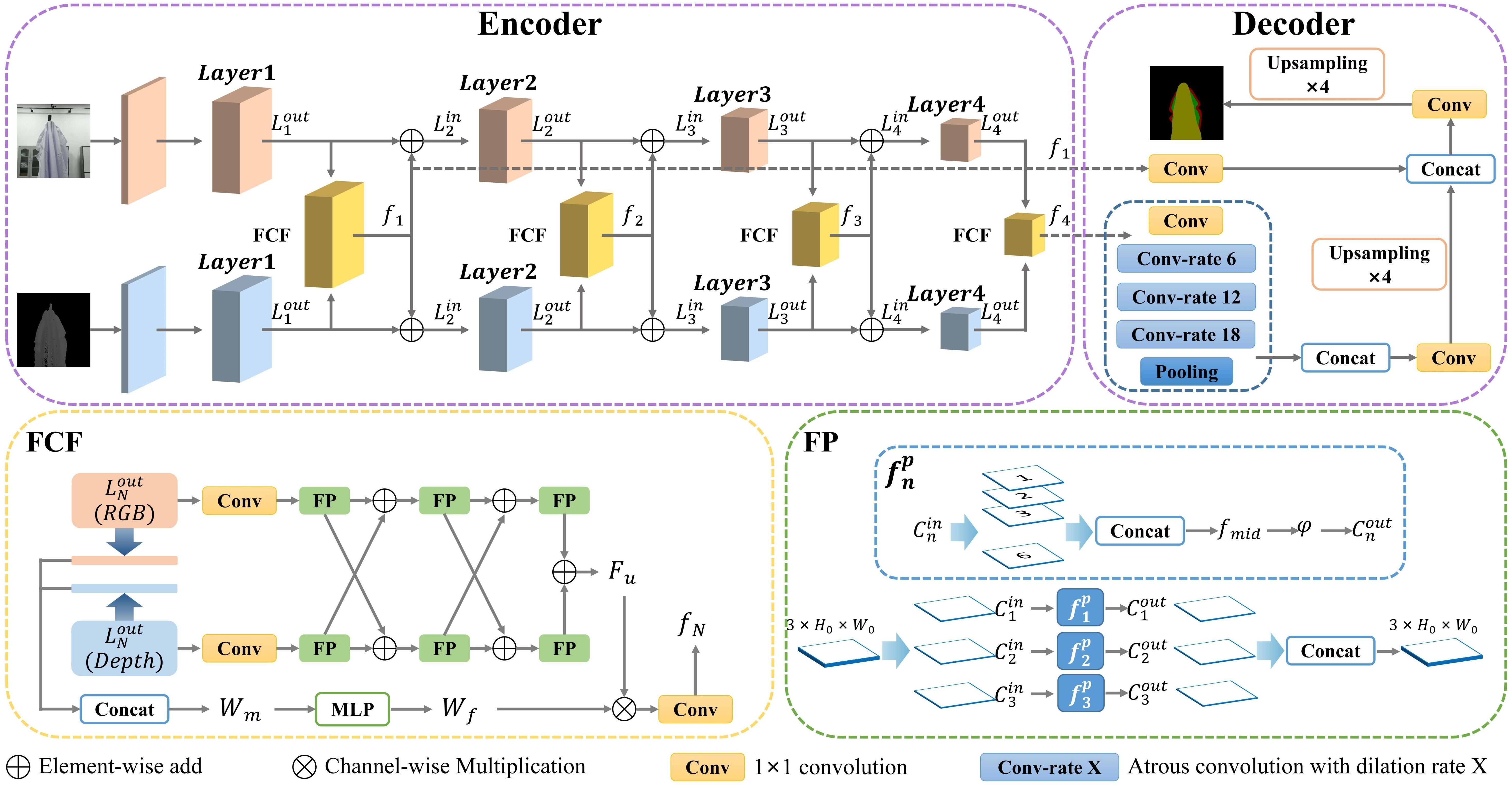}
      \caption{The overview of BiFCNet. We use a stage-by-stage Bi-directional propagation structure in the encoder. Two parallel ResNet-101 are used as the backbone and DeepLab V3+ is used as the decoder. The input of the network is a pair of RGB and depth images. Feature pairs output by each layer of ResNet-101 are fused by a FCF module and propagated to the next layer. Fusion results of the first and the last FCF modules are propagated to the DeepLab V3+.}
      \label{f3}
      \vspace{-0.5cm}
\end{figure*}
\section{RELATED WORK}
\subsection{RGB-D Semantic Segmentation}
Depth images have become practical and have been widely used in semantic segmentation tasks \cite{hazirbas2016fusenet,park2017rdfnet,zhang2019pattern,xiong2020variational,xing2020malleable,hua2019small}. At present, many RGB-D semantic segmentation models study the fusion method of RGB features and depth features, mainly considering the fusion of local features and the problem of local noise present in depth data, while ignoring the fusion of global features \cite{chen2020bi,seichter2021efficient,cao2021shapeconv,su2021deep,wang2022multimodal}. This results in poor alignment between RGB and depth features, limiting the ability of the depth's spatial information to assist in propagation of RGB semantic information. Fractal geometry has been shown to be an effective method for extracting globally complex features \cite{xu2021encoding}. In this work, we apply fractal geometry to the feature fusion process of RGB-D semantic segmentation. 
\subsection{Data Synthesis and Augmentation}
Reducing the cost of data collection and labeling while enhancing the generalization of the model is a major problem in robotics-related research. Some studies use physics engines such as Maya or Blender to synthesize large amounts of data
\cite{corona2018active,saxena2019garment,zhang2020learning,ren2021grasp,zhang2022learning,bousmalis2018using}, but the distribution of synthetic data and real data is vastly different. This leads to a requirement for domain adaptation processing \cite{ren2021grasp,zhang2022learning,bousmalis2018using}. However while this processing does reduce the domain gap, it limits the ability of the model to generalize and as a result limits the effectiveness of synthetic data for improving accuracy. Data augmentation based on adversarial strategies has been applied to a variety of vision models in the training of public datasets to enhance the generalization ability of the models \cite{cubuk2019autoaugment,suzuki2022teachaugment}. We propose a data augmentation method for the multi-input case of RGB-D semantic segmentation, and apply it to our clothing dataset training to reduce the cost of data collection and labeling, and enhance the generalization of the model.

\subsection{Robotic Manipulation of Flexible Objects}
Enabling robots to recognize and manipulate flexible objects such as clothing has been extensively studied \cite{gao2015user,li2015regrasping,gao2016iterative,tsurumine2019deep,zhang2019probabilistic,kapusta2019personalized,ha2022flingbot}. 
Many works leverage deep learning to calibrate and identify graspable points on objects \cite{ramisa2014learning,martinez2017recognition,corona2018active,saxena2019garment,zhang2020learning,jangir2020dynamic}. However, due to the high deformability of flexible objects, self-occlusion often leads to recognition failures. Current state-of-the-art research addresses this by identifying graspable regions, as opposed to keypoints, and selecting graspable points from within these regions. These methods often rely solely on depth information to perform segmentation, however this is often susceptible to noise in the depth data \cite{qian2020cloth,ren2021grasp,zhang2022learning,gabas2017physical}. Alternatively, RGB only segmentation methods struggle with single color distributions often present in clothing. Therefore, we propose an RGB-D semantic segmentation model, utilising combined RGB and depth data, to identify the graspable regions of clothing.
\begin{figure*}[!thpb]
      \centering
      \includegraphics[width=14cm]{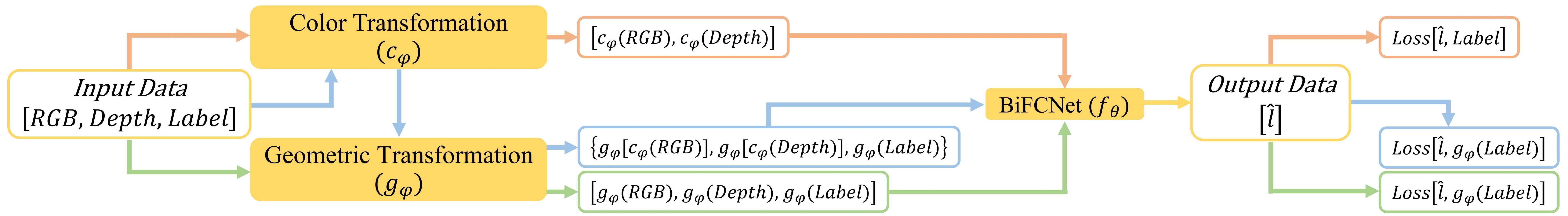}
      \caption{The process of training BiFCNet using an adversarial strategy-based data augmentation method. We build color transformation and geometric transformation based on MLP networks. The data augmentation network (color transformation and geometric transformation in the figure) and BiFCNet are alternately trained in turns to update the parameters.}
      \label{f4}
      \vspace{-0.5cm}
\end{figure*}
\section{METHODS}
\subsection{BiFCNet for RGB-D Semantic Segmentation}
Our network adopts an encoder-decoder structure, as shown in Fig \ref{f3}. The encoder is formed of two parallel ResNet-101 \cite{he2016deep} as the backbone. The output of each layer of ResNet-101 is fused in the FCF module. We set up two feature processing paths in the FCF fusion module: extracting channel weights and cross-fusion propagation. The Fractal Process (FP) module in the cross-fusion propagation path can extract global complex features based on fractal geometry. The two paths finally converge to obtain the fused feature map. We use DeepLab V3+ \cite{chen2018encoder} as the decoder. The fusion results of the first and the last FCF modules are used as the shallow feature input and deep feature input for the DeepLab V3+ network. Finally, the decoder performs feature decoding and outputs the resultant per pixel segmentation labels.

For the FCF module, RGB features and depth features are taken as inputs and denoted as $L_{n}^{out}$.
A 1x1 convolution is applied in order to change the number of channels in $L_{n}^{out}$ to n, where n is a hyperparameter, which we set it to three \cite{xu2021encoding}.
The transformed results are used as inputs to the FP module. The two types of features are cross-propagated through multiple FP modules to finally form a fusion feature $F_{u}$. At the same time, we perform a global average pooling operation on the input RGB and depth features. Two feature layers perform a stacking operation in the channel dimension to obtain $W_{m}$. We use an MLP network to process $W_{m}$ to get a set of channel weights, denoted as $W_{f}$. We perform 
channel-wise multiplication on $F_{u}$ and $W_{f}$. Finally, we apply a 1x1 convolution, to obtain a fusion feature $f_{N}$ with the same number of channels as $L_{n}^{out}$.

For the FP module, the input feature map is first separated based on the channel dimension to obtain three single-channel feature maps of $C_{1}^{in}$, $C_{2}^{in}$, and $C_{3}^{in}$.
For each $C_{n}^{in}$, $C_{n}^{out}$ is obtained through the $f_{n}^{p}$ module, which are subsequently concatenated to form the output of the FP module.
During the $f_{n}^{p}$ operation, a series of 6 convolutions with kernels of varying sizes from 1 to 6 are performed on $C_{n}^{in}$ to obtain six single-channel feature layers \cite{xu2021encoding}.
We stack the obtained six feature layers in the channel dimension to obtain feature $f_{mid}$. 
Then the process $\varphi$ of obtaining the corresponding $C_{n}^{out}$ from $f_{mid}$ can be expressed as the following formula, in which the $\overline{X}$ and $\overline{Y}$ are the results of taking the corresponding average value in the channel dimension of $X$ and $Y$. We define $F_{\oplus}$ as the sum of all channels at each location in the feature map and the $\otimes$ means channel-wise multiplication.
$$
X=\log_{2}\left[\operatorname{Relu}\left(f_{mid}\right)+1\right]\eqno{(1)}
$$
\vspace{-0.5cm}
$$
Y={tensor}\left(\log _{2} 1,\log_{2}2,\ldots,\log_{2}6\right)\eqno{(2)}
$$
\vspace{-0.5cm}
$$
C_{n}^{out}=\frac{F_{\oplus}[(X-\overline{X})\otimes(Y-\overline{Y})]}{F_{\oplus}[(Y-\overline{Y})\otimes(Y-\overline{Y})]}\eqno{(3)}
$$

\subsection{Data Augmentation Based on Adversarial Strategy}
We train BiFCNet using our proposed adversarial strategy-based data augmentation method, as shown in Figure \ref{f4}. The figure outlines three paths (red, green, and blue) showing possible data transformation situations that occur with probability $\frac{1}{4}$, $\frac{1}{2}$, and $\frac{1}{4}$, respectively. The training process for the network comprises two steps. In the first step, parameter $\varphi$ of data augmentation networks $c_{\varphi}$ and $g_{\varphi}$ is fixed, only parameter $\theta$ of BiFCNet is updated, and the loss value is minimized. In the second step, parameter $\theta$ of BiFCNet is fixed, only parameter $\varphi$ of data augmentation networks $c_{\varphi}$ and $g_{\varphi}$ is updated, and the loss value is maximized. The two steps are carried out alternately, taking the blue path as an example, the process is represented as follows.
$$
\max_{\varphi} \min_{\theta}\left\{{Loss}\left[f_{\theta}\left(g_{\varphi}\left(c_{\varphi}\right)\right),g_{\varphi}(l)\right]\right\}\eqno{(4)}
$$
\subsection{Design of Robot Operation}
\begin{figure}[!htbp]
      \centering
      \includegraphics[width=6cm]{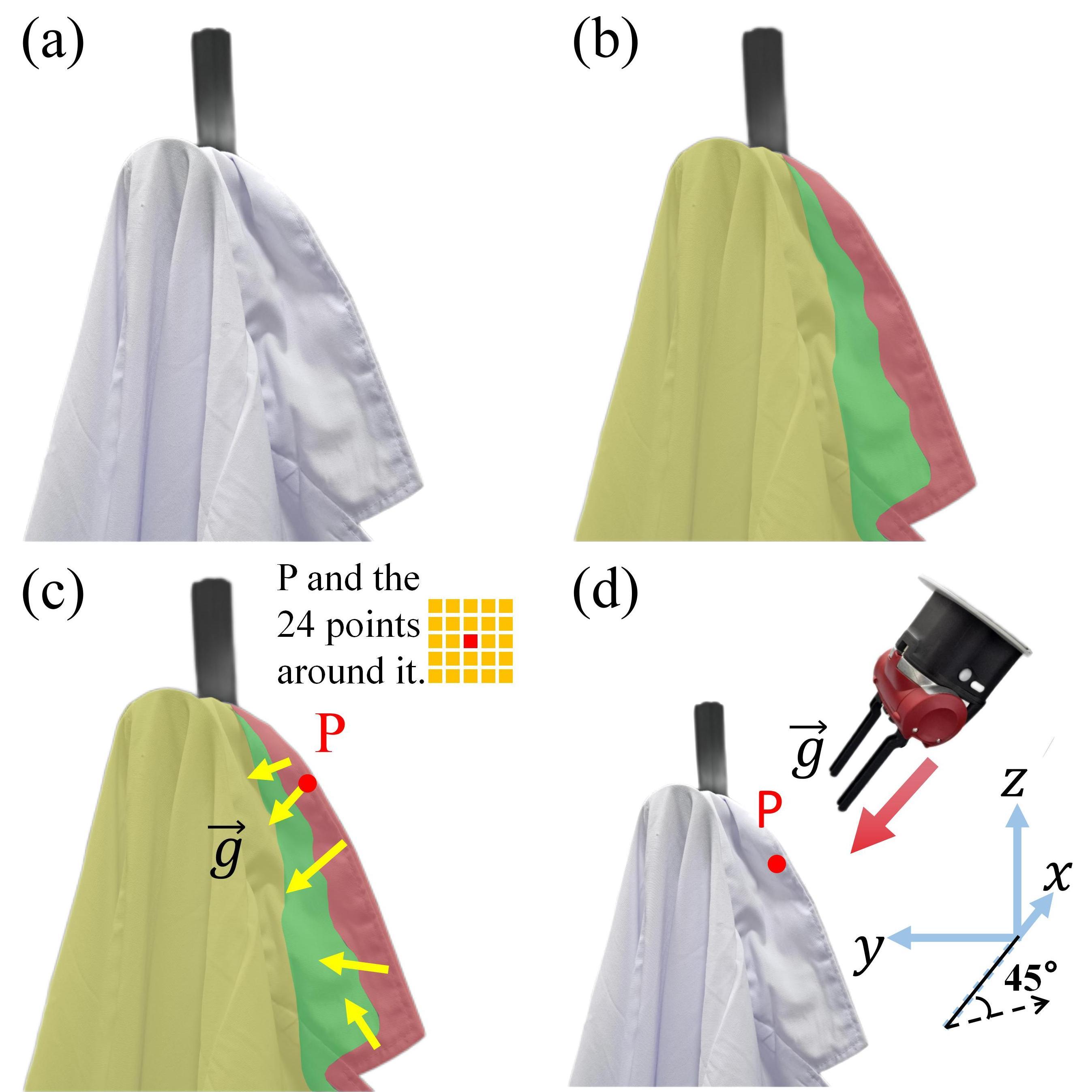}
      \caption{Strategy for selecting graspable points and grasp directions. (a) The item of clothing in the hanging state. (b) Semantic segmentation. (c) Selection of grasp point P. (d) Coordinates transformation for the calculation of grasp direction.}
      \label{f5}
      \vspace{-0.5cm}
\end{figure}

Our proposed method outputs the grasp points on the object and the corresponding grasp direction. We establish a two-dimensional plane rectangular coordinate system, where the origin lies at the geometric center of the segmented region. As shown in Figure \ref{f5}(b), for each point $d_{n}^{out}$ in the red outer edge, a point set $D_{out}$ is formed, and for each point $d_{m}^{in}$ in the green inner edge, a point set $D_{in}$ is formed.
$$
d_{n}^{out}\left(x_{n}^{out}, y_{n}^{out}\right) \in D_{out}, n=1,2,3 \ldots \eqno{(5)}
$$
\vspace{-0.5cm}
$$
d_{m}^{in}\left(x_{m}^{in}, y_{m}^{in}\right) \in D_{in}, m=1,2,3 \ldots \eqno{(6)}
$$

A direction vector $p_{n}$ is formed by searching for a point $d_{n}^{in}, d_{n}^{in} \in D_{in}$ with the smallest Euclidean distance $L_{n}^{eur}$ in the inner edge for each point $d_{n}^{out}$ in the outer edge. For each direction vector, the $sin$ and $cos$ values of the angle between it and the horizontal x-axis is calculated, denoted as $K_{n}^{sin}$ and $K_{n}^{cos}$.
$$
K_{n}^{sin}=\frac{y_{n}^{out}-y_{n}^{in}}{L_{n}^{eur}}, n=1,2,3 \ldots \eqno{(7)}
$$
$$
K_{n}^{cos}=\frac{x_{n}^{out}-x_{n}^{in}}{L_{n}^{eur}}, n=1,2,3 \ldots \eqno{(8)}
$$

For each point $d_{n}^{out}$ in the outer edge, we select the ${N}$ points around it, as shown in Figure \ref{f5}(c). The flatness $F_{n}$ of the area around this point is then represented by the result of the following formula for approximating the variance. $\overline{K_{n}^{sin}}$ and $\overline{K_{n}^{cos}}$ represent the average of $K_{n}^{sin}$ and $K_{n}^{cos}$ of these points.
$$
F_{n}=\frac{1}{N+1} \times \sum_{k=1}^{N+1}\left(K_{n}^{sin} \times K_{n}^{cos}-\overline{K_{n}^{sin}} \times \overline{K_{n}^{cos}}\right)^{2} \eqno{(9)}
$$

Finally, the point with the smallest value of $F_{n}$ is the selected graspable point, defined as point P, as shown in Figure \ref{f5}(c). We sample and average the depth values for 4 locations, each with a pixel distance of 1 around the selected point P, to determine the point P in three-dimensional space. We apply a transformation to the grasp point in order to calculate its location in the robot's coordinate system. We take the vector $\vec{g}$ of point P as the direction of the spatial plane where y-axis and the z-axis are located in the robot coordinates. We keep the projection direction of the vector $\vec{g}$ on the plane where y-axis and the z-axis are located, and then adjust the angle with the positive direction of the x-axis to 45° to get the grasping direction, as shown in Figure \ref{f5}(d).
\vspace{-0.5cm}
\section{EXPERIMENTS}
\subsection{Clothing Dataset}
\begin{figure}[t]
      \centering
      \includegraphics[width=7cm]{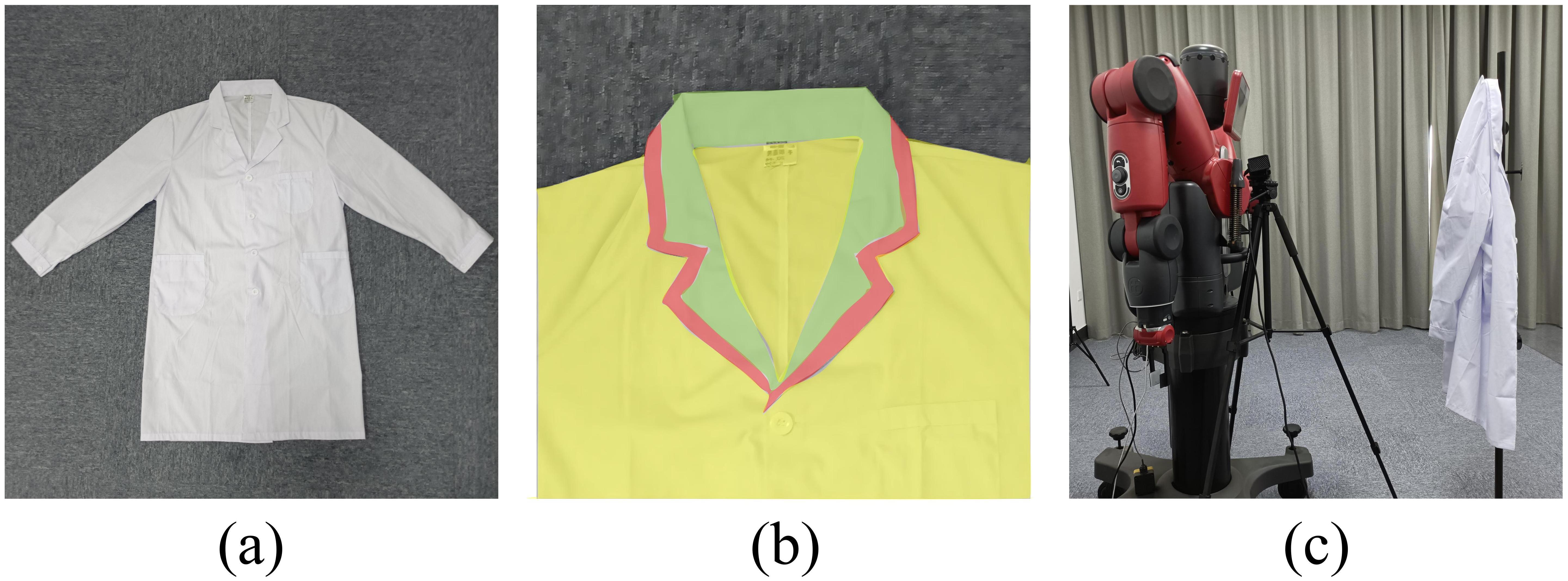}
      \caption{Collection and labeling of our clothing dataset. (a) Target clothing. (b) Semantic segmentation of the clothing item, where red represents the outer edge (the area with a width of 1.5cm from the outermost edge of the collar), green represents the inner edge (the remaining area of the collar), and yellow represents other areas of the garment. (c) Side view of the data acquisition environment.}
      \label{f6}
\end{figure}

We use a white medical suit as the target clothing, as shown in Figure \ref{f6}(a). We set up three segmentation regions, as shown in Figure \ref{f6}(b). We set up an environment for data collection in the laboratory, as shown in Figure \ref{f6}(c). This environment consists of a hanger suspended at a fixed position 0.8m in front of the robot. A Kinect camera placed in front of the robot was utilised to collect RGB-D images at a resolution of 750 by 750 pixels for both the RGB and depth components. Each time an image is collected, the hanging point of the medical clothing is manually and randomly changed to obtain a good data distribution. We collected 350 pairs of RGB-D data through this method, and used the LabelMe \cite{russell2008labelme} tool to annotate the segmented regions of the 350 RGB images. We divided the data into training and test sets consisting of 150 and 200 image pairs respectively.
\vspace{-0.5cm}
\subsection{Model Training and Evaluation}
We conduct training and evaluation experiments on NYUDv2 \cite{silberman2012indoor} and our clothing dataset, using mean Intersection-over-Union (mIoU) and pixel accuracy (PA) as two performance evaluation metric. 
We use the PyTorch framework to complete all experiments, training the model using mini-batch SGD with momentum and cross-entropy loss as the loss function. 

\begin{figure}[t]
      \centering
      \includegraphics[width=7.5cm]{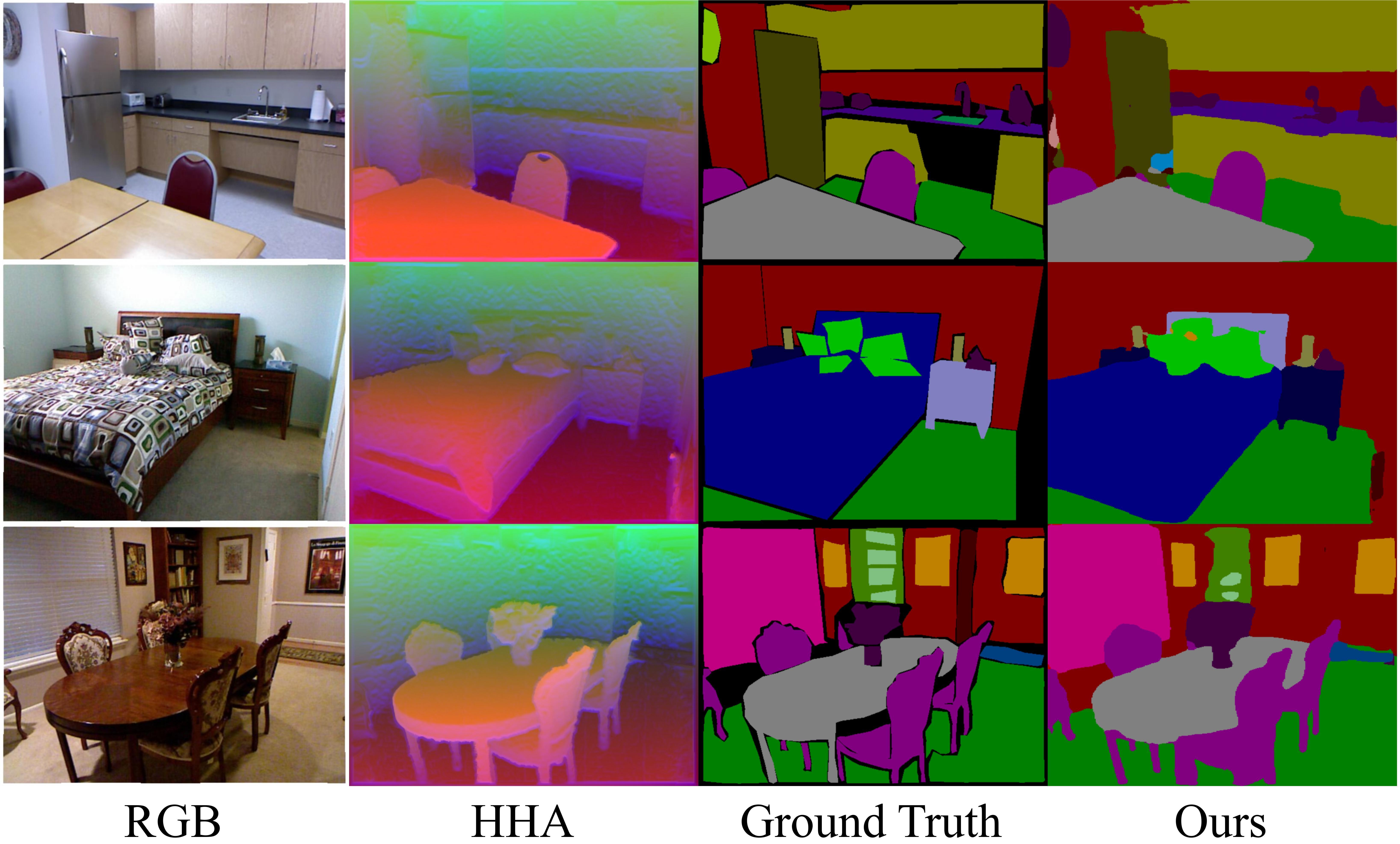}
      \caption{Visualization results on the NYUDv2.}
      \label{f7}
      \vspace{-0.25cm}
\end{figure}

For a fair comparison on NYUDv2, we choose the 40-class version and use depth images in the form of HHA \cite{gupta2014learning}, trained for 300 epochs. As shown in Table \ref{t1}, the final evaluation effect mIoU and PA on the test set reached 51.8\% and 77.9\%, respectively. Figure \ref{f7} shows the visualization results of our method.

\begin{figure}[!htbp]
      \centering
      \includegraphics[width=7.5cm]{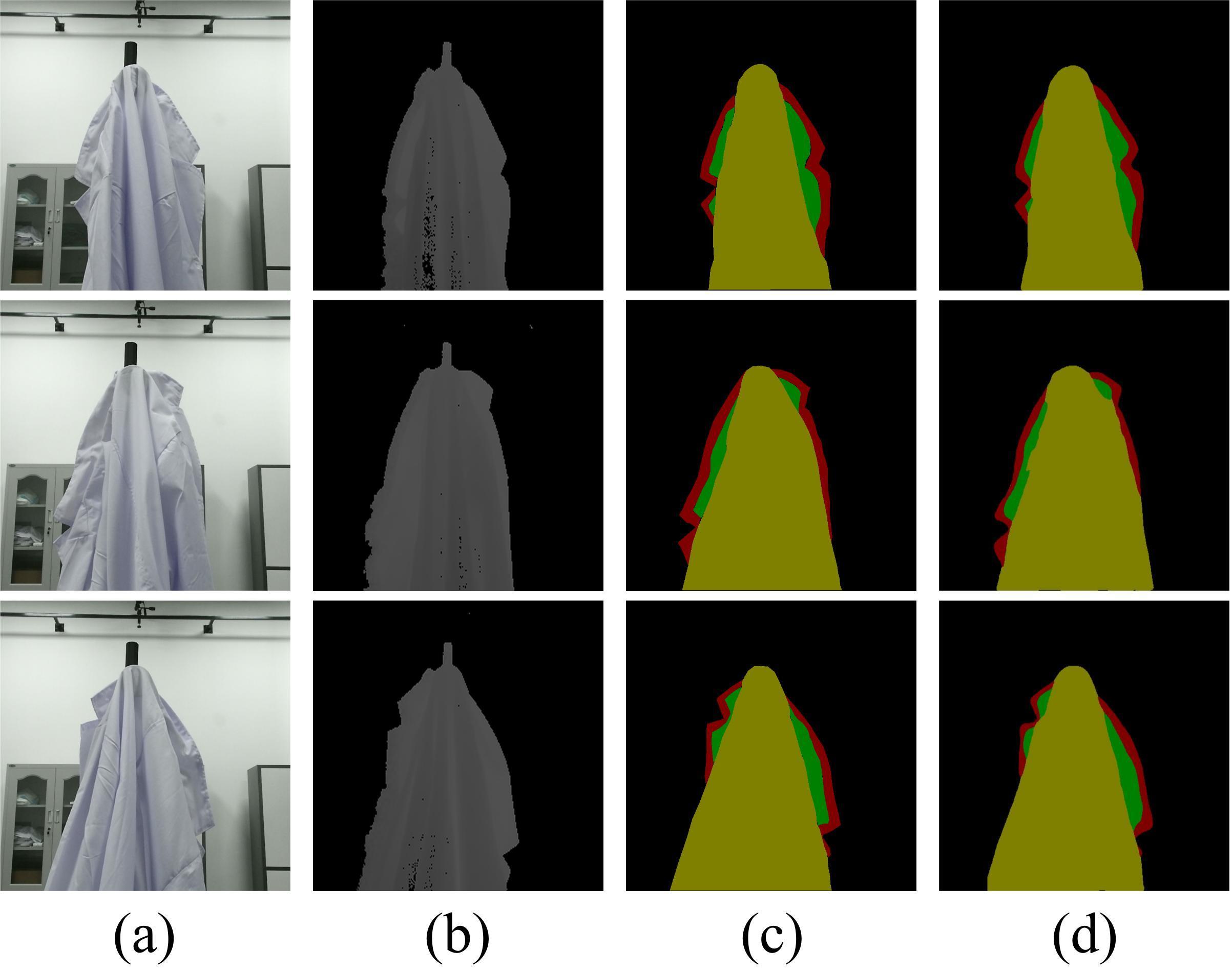}
      \caption{Visualization of the segmentation results on the clothing dataset. (a) RGB images of the target garment. (b) Depth images of the target clothing. (c) Manually annotated label images. (d) Segmentation results using our proposed BiFCNet.}
      \label{f8}
      \vspace{-0.65cm}
\end{figure}

For the clothing dataset, we set 4 classes, namely Background, Outer Edge, Inner Edge and Other Parts of Clothing, and trained for 60 epochs. As shown in Table \ref{t2}, the final evaluation results mIoU and PA on the test set reached 85.13\% and 97.95\%, respectively. Figure \ref{f8} shows the visualization results of our method.

\begin{table}[t]
\caption{Experiment results compared with state-of-the-art RGB-D semantic segmentation models on NYUDv2.}
\begin{center}
\begin{tabular}{cccc}
\toprule
Method & Backbone & mIoU $\uparrow$ & PA $\uparrow$\\
\midrule
ESANet\cite{seichter2021efficient} & ResNet-50 & 50.5\% & / \\
VCD+RedNet\cite{xiong2020variational} & ResNet-50 & 50.7\% & / \\
Malleable 2.5D\cite{xing2020malleable} & ResNet-101 & 50.9\% & 76.9\%\\
ShapeConv\cite{cao2021shapeconv} & ResNet-101 & 51.3\% & 76.4\%\\
CANet\cite{zhou2022canet} & ResNet-101 & 51.5\% & 77.1\%\\
BiFCNet (Ours) & ResNet-101 & \textbf{51.8\%} & \textbf{77.9\%}\\
\bottomrule
\end{tabular}
\end{center}
\label{t1}
\vspace{-0.5cm}
\end{table}

\begin{table}[!htbp]
\caption{Segmentation results on the clothing dataset. We note that while ``Other Parts of Clothing'' and ``Background'' achieves IoU values of over 97\%, ``Inner Edge'' and ``Outer Edge'' only achieve 70\%. This is due to the relatively small areas and complicated shapes of inner and out collar.}

\begin{center}
\begin{tabular}{cccc}
\toprule
Classes & IoU $\uparrow$ & mIoU $\uparrow$ & PA $\uparrow$\\
\midrule
Background & 97.45\% & \multirow{4}*{85.13\%} & \multirow{4}*{97.95\%}\\
Outer Edge & 74.65\% & ~ & ~\\
Inner Edge & 70.36\% & ~ & ~\\
Other Parts of Clothing & 98.08\% & ~ & ~\\
\bottomrule
\end{tabular}
\end{center}
\label{t2}
\end{table}

\subsection{Evaluation of Robot Operation}
We deploy the model trained on the clothing dataset to the Baxter system.
Input RGB and depth images are collected by the Kinect camera placed in front of the robot. These images are then passed through our proposed BiFCNet producing corresponding segmentation maps.
According to the designed point selection strategy, the graspable points on both sides are determined, and the coordinates are transformed into robot coordinates with respect to the depth information. In relation to the calculated grasp points and grasp direction, a path is planned, and the robot starts the grasping and unfolding operation. The success criteria for a successful grasp and unfold, requires the robots two grippers to successfully grasp the item of clothing, remove the item from the hanger and perform the unfold operation. We randomly changed the hanging point of the clothes and conducted 100 experiments. The success rate reached 84\%, as shown in the first row of Table \ref{t3}, in the third row of Table \ref{t4} and in the first row of Table \ref{t5}.
 
\subsection{Ablation Study}
We conducted a series of ablation experiments to evaluate our method. For each trained model, we completed 100 grasping and unfolding tasks on the Baxter robotic system.


1) To demonstrate the effectiveness of the data augmentation method, we evaluate the performance of the proposed pipeline with and without our data augmentation component on varying quantities of training data. The experimental results are shown in Table \ref{t3}. 
We first tested the use of our proposed method with and without data augmentation on a dataset of 150 training image pairs. Additionally, we collected and annotated 300 pairs of RGB-D images, compared the effects of dataset size by training a model without data augmentation on the total 450 training image pairs. For a fair comparison, we trained the models with a dataset size of 150 images for 60 epochs, where as the model with a dataset size of 450 images for only 20 epochs.
Our data augmentation method results in a higher percentage of successful grasps and unfolds outperforming both cases where data augmentation is not used, even when additional training data is provided. This shows that data augmentation is effective for improving model generalization, improving the generalization of the model beyond the method of manual collection of additional annotated real data.
\begin{table}[!htbp]
\caption{Results of ablation experiments considering data scale and data augmentation factors. We note that the performance drops significantly without the use of data augmentation, even when dataset size is accounted for.
}
\vspace{-1em}
\begin{center}
\begin{tabular}{ccccc}
\toprule
Train & Data & \multirow{2}*{mIoU $\uparrow$} & \multirow{2}*{PA $\uparrow$} & Grasp and Unfold\\
 Scale & Augmentation &  &  & Success Rate $\uparrow$\\
\midrule
150 & \checkmark & 85.13\% & 97.95\% & \textbf{84\%}\\
150 & $\times$ & 65.25\% & 79.68\% & 57\%\\
450 & $\times$ & 76.18\% & 95.67\% & 74\%\\
\bottomrule
\end{tabular}
\end{center}
\label{t3}
\vspace{-0.4cm}
\end{table}

2) To demonstrate the effectiveness of the proposed BiFCNet, we evaluate our pipeline with and without the FCF feature fusion method by varying the input data types.
We first evaluated the pipeline with segmentation only performed on RGB data as a baseline, using DeepLab V3+ \cite{chen2018encoder} as the semantic segmentation model and ResNet-101 as the backbone. We compared this with two methods. First the use of RGB-D data where features are fused via stacking (two parallel ResNet-101 networks are used as the backbone) and finally the use of RGB-D data with our full FCF pipeline. 
As shown in Table \ref{t4}, the experiments show that use of both RGB and depth information is advantageous compared to use of RGB information only. These results further show that our FCF module is able to outperform stacking to fuse RGB and depth data. This is due to the stacking method not considering the extraction of global features.
\begin{table}[!htbp]
\caption{Results of ablation experiments considering different types of input data and different fusion methods. The results show that the use of our FCF module is advantageous to the use of stacking as a fusion method.}
\vspace{-1em}
\begin{center}
\begin{tabular}{ccccc}
\toprule
\multirow{2}*{Inputs} & Feature Fusion & \multirow{2}*{mIoU $\uparrow$} & \multirow{2}*{PA $\uparrow$} & Grasp and Unfold\\
  & Method &  &  & Success Rate $\uparrow$\\
\midrule
RGB & None & 76.98\% & 96.8\% & 73\%\\
RGB-D & Stacking & 81.21\% & 97.35\% & 79\%\\
RGB-D & FCF & 85.13\% & 97.95\% & \textbf{84\%}\\
\bottomrule
\end{tabular}
\end{center}
\label{t4}
\vspace{-0.4cm}
\end{table}

3) In order to demonstrate the effectiveness of the proposed method for selecting grasping points from the graspable regions, we evaluate the performance of our method compared with a strategy of random point selection from the the outer edge region of the clothing item. 
We additionally compare our method against a method based on ResNet-101 which directly regresses to graspable points instead of identifying graspable regions first. This final method utilises only depth data as input.
The experimental results are shown in Table \ref{t5}.
These results demonstrate how first identifying graspable regions before identifying graspable points, results in higher quality grasp point prediction.
\begin{table}[!htbp]
\caption{Results of ablation experiments considering different graspable point selection methods. Our proposed method outperforms both random selection and direct identification of grasp points without segmentation.}
\vspace{-1em}
\begin{center}
\begin{tabular}{ccccc}
\toprule
Recognition & Point & \multirow{2}*{mIoU $\uparrow$} & \multirow{2}*{PA $\uparrow$} & Grasp and Unfold\\
Method & Selection &  &  & Success Rate $\uparrow$\\
\midrule
Ours & Ours & 85.13\% & 97.95\% & \textbf{84\%}\\
Ours & Random & 85.13\% & 97.95\% & 57\%\\
ResNet-101 & / & / & / & 61\%\\
\bottomrule
\end{tabular}
\end{center}
\label{t5}
\vspace{-0.4cm}
\end{table}

\section{CONCLUSIONS}
In this paper, we propose an RGB-D semantic segmentation network BiFCNet to allow robots to recognize graspable regions of clothing. The FCF module in our network can consider global complex features based on fractal geometry to achieve fusion of RGB and depth data. We also propose an adversarial strategy-based data augmentation method to train BiFCNet, thereby reducing the cost of real data collection. A pipeline for clothes grasping and unfolding is proposed, by combining BiFCNet with a strategy to select graspable points from segmentation regions based on clothing flatness measures while taking into account the grasping direction. We evaluate BiFCNet on the public dataset NYUDv2, achieving comparable results to the state-of-the-art models. We deploy our model on Baxter for extensive grasping and unfolding experiments to demonstrate the effectiveness and superiority of our proposed method. In the future, we will consider unsupervised methods for semantic segmentation to further reduce the cost of data annotation. 

\section*{ACKNOWLEDGMENT}


This work is supported in part by the National Natural Science Foundation of China under grant No. 62203184 and the International Cooperation Project under grant No. 20220402009GH. This work is also supported in part by the MSIT, Korea, under the ITRC program (IITP-2022-2020-0-01789) (50\%) and the High-Potential Individuals Global Training Program (RS-2022-00155054) (50\%) supervised by the IITP.

\clearpage

\bibliographystyle{ieeetr} 
\bibliography{ref}

\end{document}

%% file: macro.tex
\usepackage{color}
\usepackage[normalem]{ulem}
\definecolor{turquoise}{cmyk}{0.65,0,0.1,0.3}
\definecolor{purple}{rgb}{0.65,0,0.65}
\definecolor{dark_green}{rgb}{0, 0.5, 0}
\definecolor{orange}{rgb}{1.0, 0.27, 0.0}
\definecolor{red}{rgb}{0.8, 0.2, 0.2}
\definecolor{darkred}{rgb}{0.6, 0.1, 0.05}
\definecolor{blueish}{rgb}{0.0, 0.3, .6}
\definecolor{light_gray}{rgb}{0.7, 0.7, .7}
\definecolor{pink}{rgb}{1, 0, 1}
\definecolor{greyblue}{rgb}{0.25, 0.25, 1}

%% file: root.bbl
\begin{thebibliography}{10}

\bibitem{zhang2022learning}
F.~Zhang and Y.~Demiris, ``Learning garment manipulation policies toward
  robot-assisted dressing,'' {\em Science robotics}, vol.~7, no.~65,
  p.~eabm6010, 2022.

\bibitem{gao2015user}
Y.~Gao, H.~J. Chang, and Y.~Demiris, ``User modelling for personalised dressing
  assistance by humanoid robots,'' in {\em IEEE/RSJ International Conference on
  Intelligent Robots and Systems (IROS)}, pp.~1840--1845, 2015.

\bibitem{zhang2019probabilistic}
F.~Zhang, A.~Cully, and Y.~Demiris, ``Probabilistic real-time user posture
  tracking for personalized robot-assisted dressing,'' {\em IEEE Transactions
  on Robotics}, vol.~35, no.~4, pp.~873--888, 2019.

\bibitem{gao2016iterative}
Y.~Gao, H.~J. Chang, and Y.~Demiris, ``Iterative path optimisation for
  personalised dressing assistance using vision and force information,'' in
  {\em IEEE/RSJ international conference on intelligent robots and systems
  (IROS)}, pp.~4398--4403, 2016.

\bibitem{zhang2020learning}
F.~Zhang and Y.~Demiris, ``Learning grasping points for garment manipulation in
  robot-assisted dressing,'' in {\em IEEE International Conference on Robotics
  and Automation (ICRA)}, pp.~9114--9120, 2020.

\bibitem{saxena2019garment}
K.~Saxena and T.~Shibata, ``Garment recognition and grasping point detection
  for clothing assistance task using deep learning,'' in {\em IEEE/SICE
  International Symposium on System Integration (SII)}, pp.~632--637, 2019.

\bibitem{gao2020user}
Y.~Gao, H.~J. Chang, and Y.~Demiris, ``User modelling using multimodal
  information for personalised dressing assistance,'' {\em IEEE Access},
  vol.~8, pp.~45700--45714, 2020.

\bibitem{ren2021grasp}
R.~Ren, M.~G. Rajesh, J.~Sanchez-Riera, F.~Zhang, Y.~Tian, A.~Agudo,
  Y.~Demiris, K.~Mikolajczyk, and F.~Moreno-Noguer, ``Grasp-oriented
  fine-grained cloth segmentation without real supervision,'' {\em arXiv
  preprint arXiv:2110.02903}, 2021.

\bibitem{bousmalis2018using}
K.~Bousmalis, A.~Irpan, P.~Wohlhart, Y.~Bai, M.~Kelcey, M.~Kalakrishnan,
  L.~Downs, J.~Ibarz, P.~Pastor, K.~Konolige, {\em et~al.}, ``Using simulation
  and domain adaptation to improve efficiency of deep robotic grasping,'' in
  {\em IEEE international conference on robotics and automation (ICRA)},
  pp.~4243--4250, 2018.

\bibitem{hazirbas2016fusenet}
C.~Hazirbas, L.~Ma, C.~Domokos, and D.~Cremers, ``Fusenet: Incorporating depth
  into semantic segmentation via fusion-based cnn architecture,'' in {\em Asian
  conference on computer vision}, pp.~213--228, Springer, 2016.

\bibitem{park2017rdfnet}
S.-J. Park, K.-S. Hong, and S.~Lee, ``Rdfnet: Rgb-d multi-level residual
  feature fusion for indoor semantic segmentation,'' in {\em Proceedings of the
  IEEE international conference on computer vision}, pp.~4980--4989, 2017.

\bibitem{zhang2019pattern}
Z.~Zhang, Z.~Cui, C.~Xu, Y.~Yan, N.~Sebe, and J.~Yang, ``Pattern-affinitive
  propagation across depth, surface normal and semantic segmentation,'' in {\em
  Proceedings of the IEEE/CVF conference on computer vision and pattern
  recognition}, pp.~4106--4115, 2019.

\bibitem{xiong2020variational}
Z.~Xiong, Y.~Yuan, N.~Guo, and Q.~Wang, ``Variational context-deformable
  convnets for indoor scene parsing,'' in {\em Proceedings of the IEEE/CVF
  Conference on Computer Vision and Pattern Recognition}, pp.~3992--4002, 2020.

\bibitem{xing2020malleable}
Y.~Xing, J.~Wang, and G.~Zeng, ``Malleable 2.5 d convolution: Learning
  receptive fields along the depth-axis for rgb-d scene parsing,'' in {\em
  European Conference on Computer Vision}, pp.~555--571, Springer, 2020.

\bibitem{hua2019small}
M.~Hua, Y.~Nan, and S.~Lian, ``Small obstacle avoidance based on rgb-d semantic
  segmentation,'' in {\em Proceedings of the IEEE/CVF International Conference
  on Computer Vision Workshops}, pp.~0--0, 2019.

\bibitem{schwarz2018rgb}
M.~Schwarz, A.~Milan, A.~S. Periyasamy, and S.~Behnke, ``Rgb-d object detection
  and semantic segmentation for autonomous manipulation in clutter,'' {\em The
  International Journal of Robotics Research}, vol.~37, no.~4-5, pp.~437--451,
  2018.

\bibitem{chen2020bi}
X.~Chen, K.-Y. Lin, J.~Wang, W.~Wu, C.~Qian, H.~Li, and G.~Zeng,
  ``Bi-directional cross-modality feature propagation with
  separation-and-aggregation gate for rgb-d semantic segmentation,'' in {\em
  European Conference on Computer Vision}, pp.~561--577, Springer, 2020.

\bibitem{seichter2021efficient}
D.~Seichter, M.~K{\"o}hler, B.~Lewandowski, T.~Wengefeld, and H.-M. Gross,
  ``Efficient rgb-d semantic segmentation for indoor scene analysis,'' in {\em
  IEEE International Conference on Robotics and Automation (ICRA)},
  pp.~13525--13531, 2021.

\bibitem{cao2021shapeconv}
J.~Cao, H.~Leng, D.~Lischinski, D.~Cohen-Or, C.~Tu, and Y.~Li, ``Shapeconv:
  Shape-aware convolutional layer for indoor rgb-d semantic segmentation,'' in
  {\em Proceedings of the IEEE/CVF International Conference on Computer
  Vision}, pp.~7088--7097, 2021.

\bibitem{su2021deep}
Y.~Su, Y.~Yuan, and Z.~Jiang, ``Deep feature selection-and-fusion for rgb-d
  semantic segmentation,'' in {\em IEEE International Conference on Multimedia
  and Expo (ICME)}, pp.~1--6, 2021.

\bibitem{wang2022multimodal}
Y.~Wang, X.~Chen, L.~Cao, W.~Huang, F.~Sun, and Y.~Wang, ``Multimodal token
  fusion for vision transformers,'' in {\em Proceedings of the IEEE/CVF
  Conference on Computer Vision and Pattern Recognition}, pp.~12186--12195,
  2022.

\bibitem{xu2021encoding}
Y.~Xu, F.~Li, Z.~Chen, J.~Liang, and Y.~Quan, ``Encoding spatial distribution
  of convolutional features for texture representation,'' {\em Advances in
  Neural Information Processing Systems}, vol.~34, pp.~22732--22744, 2021.

\bibitem{cubuk2019autoaugment}
E.~D. Cubuk, B.~Zoph, D.~Mane, V.~Vasudevan, and Q.~V. Le, ``Autoaugment:
  Learning augmentation strategies from data,'' in {\em Proceedings of the
  IEEE/CVF Conference on Computer Vision and Pattern Recognition},
  pp.~113--123, 2019.

\bibitem{suzuki2022teachaugment}
T.~Suzuki, ``Teachaugment: Data augmentation optimization using teacher
  knowledge,'' in {\em Proceedings of the IEEE/CVF Conference on Computer
  Vision and Pattern Recognition}, pp.~10904--10914, 2022.

\bibitem{corona2018active}
E.~Corona, G.~Alenya, A.~Gabas, and C.~Torras, ``Active garment recognition and
  target grasping point detection using deep learning,'' {\em Pattern
  Recognition}, vol.~74, pp.~629--641, 2018.

\bibitem{li2015regrasping}
Y.~Li, D.~Xu, Y.~Yue, Y.~Wang, S.-F. Chang, E.~Grinspun, and P.~K. Allen,
  ``Regrasping and unfolding of garments using predictive thin shell
  modeling,'' in {\em IEEE International Conference on Robotics and Automation
  (ICRA)}, pp.~1382--1388, 2015.

\bibitem{tsurumine2019deep}
Y.~Tsurumine, Y.~Cui, E.~Uchibe, and T.~Matsubara, ``Deep reinforcement
  learning with smooth policy update: Application to robotic cloth
  manipulation,'' {\em Robotics and Autonomous Systems}, vol.~112, pp.~72--83,
  2019.

\bibitem{kapusta2019personalized}
A.~Kapusta, Z.~Erickson, H.~M. Clever, W.~Yu, C.~K. Liu, G.~Turk, and C.~C.
  Kemp, ``Personalized collaborative plans for robot-assisted dressing via
  optimization and simulation,'' {\em Autonomous Robots}, vol.~43, no.~8,
  pp.~2183--2207, 2019.

\bibitem{ha2022flingbot}
H.~Ha and S.~Song, ``Flingbot: The unreasonable effectiveness of dynamic
  manipulation for cloth unfolding,'' in {\em Conference on Robot Learning},
  pp.~24--33, PMLR, 2022.

\bibitem{ramisa2014learning}
A.~Ramisa, G.~Alenya, F.~Moreno-Noguer, and C.~Torras, ``Learning rgb-d
  descriptors of garment parts for informed robot grasping,'' {\em Engineering
  Applications of Artificial Intelligence}, vol.~35, pp.~246--258, 2014.

\bibitem{martinez2017recognition}
L.~M. Mart{\'\i}nez and J.~Ruiz-del Solar, ``Recognition of grasp points for
  clothes manipulation under unconstrained conditions,'' in {\em Robot World
  Cup}, pp.~350--362, Springer, 2017.

\bibitem{jangir2020dynamic}
R.~Jangir, G.~Alenya, and C.~Torras, ``Dynamic cloth manipulation with deep
  reinforcement learning,'' in {\em IEEE International Conference on Robotics
  and Automation (ICRA)}, pp.~4630--4636, 2020.

\bibitem{qian2020cloth}
J.~Qian, T.~Weng, L.~Zhang, B.~Okorn, and D.~Held, ``Cloth region segmentation
  for robust grasp selection,'' in {\em IEEE/RSJ International Conference on
  Intelligent Robots and Systems (IROS)}, pp.~9553--9560, 2020.

\bibitem{gabas2017physical}
A.~Gabas and Y.~Kita, ``Physical edge detection in clothing items for robotic
  manipulation,'' in {\em International Conference on Advanced Robotics
  (ICAR)}, pp.~524--529, 2017.

\bibitem{he2016deep}
K.~He, X.~Zhang, S.~Ren, and J.~Sun, ``Deep residual learning for image
  recognition,'' in {\em Proceedings of the IEEE conference on computer vision
  and pattern recognition}, pp.~770--778, 2016.

\bibitem{chen2018encoder}
L.-C. Chen, Y.~Zhu, G.~Papandreou, F.~Schroff, and H.~Adam, ``Encoder-decoder
  with atrous separable convolution for semantic image segmentation,'' in {\em
  Proceedings of the European conference on computer vision (ECCV)},
  pp.~801--818, 2018.

\bibitem{russell2008labelme}
B.~C. Russell, A.~Torralba, K.~P. Murphy, and W.~T. Freeman, ``Labelme: a
  database and web-based tool for image annotation,'' {\em International
  journal of computer vision}, vol.~77, no.~1, pp.~157--173, 2008.

\bibitem{silberman2012indoor}
N.~Silberman, D.~Hoiem, P.~Kohli, and R.~Fergus, ``Indoor segmentation and
  support inference from rgbd images,'' in {\em European conference on computer
  vision}, pp.~746--760, Springer, 2012.

\bibitem{gupta2014learning}
S.~Gupta, R.~Girshick, P.~Arbel{\'a}ez, and J.~Malik, ``Learning rich features
  from rgb-d images for object detection and segmentation,'' in {\em European
  conference on computer vision}, pp.~345--360, Springer, 2014.

\bibitem{zhou2022canet}
H.~Zhou, L.~Qi, H.~Huang, X.~Yang, Z.~Wan, and X.~Wen, ``Canet: Co-attention
  network for rgb-d semantic segmentation,'' {\em Pattern Recognition},
  vol.~124, p.~108468, 2022.

\end{thebibliography}
